\pdfoutput=1
\documentclass[11pt]{article}

\usepackage[utf8]{inputenc}
\usepackage[T1]{fontenc}
\usepackage[margin=1in]{geometry}
\usepackage[numbers,sort&compress]{natbib}
\usepackage{amssymb}
\usepackage{amsmath}
\usepackage{booktabs}
\usepackage[hidelinks,hypertexnames=false]{hyperref}
\usepackage{graphicx}
\graphicspath{{./}}
\usepackage{multirow}
\usepackage{algorithm}
\usepackage{algpseudocode}
\usepackage{xcolor}
\algnewcommand{\algorithmicinput}{\textbf{Input:}}
\algnewcommand{\Input}{\item[\algorithmicinput]}
\algnewcommand{\algorithmicoutput}{\textbf{Output:}}
\algnewcommand{\Output}{\item[\algorithmicoutput]}
\newcommand{\proc}[1]{\text{\textsc{#1}}}

\title{SpaceMind: A Modular and Self-Evolving Embodied Vision-Language Agent Framework for Autonomous On-orbit Servicing}

\author{
  Aodi Wu$^{1,2}$ \and
  Haodong Han$^{1,2}$ \and
  Xubo Luo$^{1,2}$ \and
  Ruisuo Wang$^{2}$ \and
  Shan He$^{2}$ \and
  Xue Wan$^{2}$\thanks{Corresponding author. Email: wuaodi20@mails.ucas.ac.cn (Aodi Wu), wanxue@csu.ac.cn (Xue Wan).}
}

\date{
  $^1$University of Chinese Academy of Sciences, Beijing, China\\
  $^2$Technology and Engineering Center for Space Utilization, Chinese Academy of Sciences, Beijing, China
}

\begin{document}
\maketitle

\begin{abstract}
Autonomous on-orbit servicing demands embodied agents that perceive through visual sensors, reason about 3D spatial situations, and execute multi-phase tasks over extended horizons. We present SpaceMind, a modular and self-evolving vision-language model (VLM) agent framework that decomposes knowledge, tools, and reasoning into three independently extensible dimensions: skill modules with dynamic routing, Model Context Protocol (MCP) tools with configurable profiles, and injectable reasoning-mode skills. An MCP-Redis interface layer enables the same codebase to operate across simulation and physical hardware without modification, and a Skill Self-Evolution mechanism distills operational experience into persistent skill files without model fine-tuning. We validate SpaceMind through 192 closed-loop runs across five satellites, three task types, and two environments, a UE5 simulation and a physical laboratory, deliberately including degraded conditions to stress-test robustness. Under nominal conditions all modes achieve 90--100\% navigation success; under degradation, the Prospective mode uniquely succeeds in search-and-approach tasks where other modes fail. A self-evolution study shows that the agent recovers from failure in four of six groups from a single failed episode, including complete failure to 100\% success and inspection scores improving from 12 to 59 out of 100. Real-world validation confirms zero-code-modification transfer to a physical robot with 100\% rendezvous success. Code: \url{https://github.com/wuaodi/SpaceMind}.
\end{abstract}

\section{Introduction}
\label{sec:intro}

\begin{figure}[!t]
  \centering
  \includegraphics[width=\linewidth]{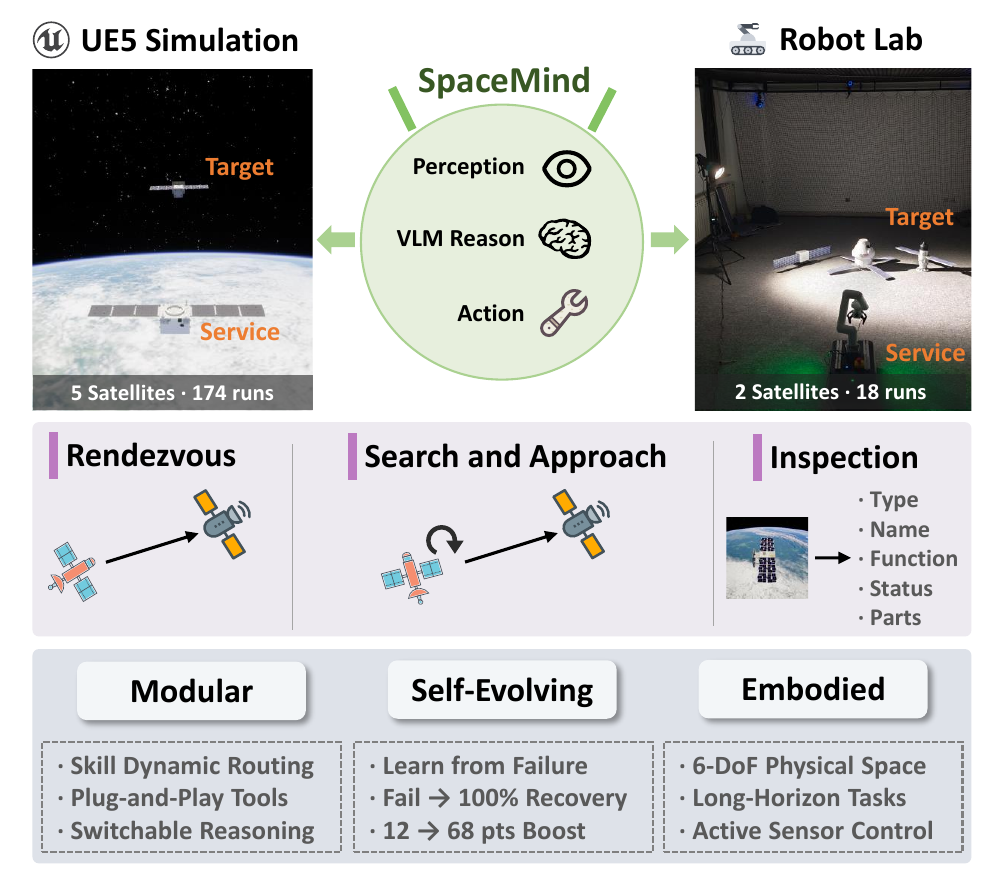}
  \caption{Overview of SpaceMind. The framework operates as a VLM-based decision-control hub that perceives through visual sensors, reasons about the current situation, and issues motion and sensor-control commands. It supports three task types, namely rendezvous, search-and-approach, and inspection, across both high-fidelity UE5 simulation with 5~satellites and 174~runs and a physical laboratory with 2~satellites and 18~runs. Key capabilities include modular skill routing, plug-and-play tools, switchable reasoning modes, and experience-driven self-evolution.}
  \label{fig:teaser}
\end{figure}

The growing population of space debris and aging satellites in low Earth orbit, which exceeded 20{,}000 cataloged objects as of 2023~\cite{nasa2023odqn}, has made on-orbit servicing, assembly, and manufacturing (OSAM) an increasingly critical capability~\cite{arney2021osam,flores2014review}. Missions such as NASA's OSAM-2~\cite{nasa2023osam2}, ESA's ClearSpace-1~\cite{biesbroek2021clearspace}, and RemoveDEBRIS~\cite{forshaw2016removedebris} have demonstrated early concepts for active debris removal and satellite life extension. However, current operational approaches rely on pre-programmed sequences designed for known cooperative targets, limiting their ability to handle the diverse and unpredictable conditions of real on-orbit encounters. Recent studies have begun exploring vision-language models (VLMs) as spacecraft operator agents~\cite{rodriguez2024lmspacecraft,carrasco2025vlm}, suggesting that large foundation models can decompose complex space tasks through natural-language reasoning. Yet these efforts remain at the demonstration level, operating in simplified game environments by reading screen text rather than processing actual visual sensor data, addressing only single tasks with monolithic prompts, and lacking systematic evaluation across targets or environments. More broadly, the current AI agent research landscape~\cite{wang2024llmagent} is dominated by digital-world applications such as web navigation, code generation, and dialogue systems, or simple physical scenarios such as tabletop manipulation and indoor navigation. Autonomous on-orbit servicing, by contrast, demands \emph{embodied} agents that perceive through real sensors, execute 6-degree-of-freedom (DoF) motion control in a physical 3D environment, and sustain multi-phase tasks spanning dozens of decision steps, from long-range search to close-range approach to fine-grained diagnosis (Fig.~\ref{fig:teaser}). How to build a scalable, adaptable, and sensor-driven agent framework for this domain remains an open problem.

Building such a framework entails four interrelated challenges. \emph{First}, there is no modular agent architecture designed for space operations. Existing space agent prototypes encode all task knowledge in a single monolithic prompt, which becomes unmaintainable as the number of tasks grows; simulation and real hardware typically require separate codebases, with no interface abstraction to bridge them. \emph{Second}, space agents lack flexible reasoning depth and cross-step memory. Different task complexities call for different decision depths; a straightforward rendezvous may require only direct action selection, whereas a search-and-approach task benefits from deliberative multi-candidate evaluation. Yet current approaches offer no mechanism to switch reasoning strategies. Moreover, agents need persistent memory across decision steps to recover from target loss and dynamically adjust strategies, which existing space agent designs do not support. \emph{Third}, space agents cannot improve from experience. Current designs are one-shot executors that start each episode from scratch, forgoing the operational knowledge gained in prior runs. Space operations are inherently repeated-execution scenarios where the same task types recur across different targets and conditions; an agent that accumulates experience could substantially improve long-term reliability. Traditional adaptation methods are impractical for on-orbit deployment: fine-tuning requires labeled data and gradient updates, reinforcement learning demands explicit reward function design, and in-context learning is bounded by the context window and does not persist across sessions. \emph{Fourth}, the design decisions behind space agent architectures lack systematic empirical validation. Prior studies~\cite{rodriguez2024lmspacecraft,carrasco2025vlm} report results on a single task with a handful of runs, providing no evidence on whether modular architectures scale, which reasoning mode to select, whether self-evolution is effective, or whether the same architecture can transfer across environments.

To address these challenges, we present SpaceMind, a modular and self-evolving embodied VLM-agent framework for autonomous on-orbit servicing (Fig.~\ref{fig:architecture}). SpaceMind uses a VLM as a decision-control hub that perceives the environment through visual sensors, reasons about the current situation, invokes specialized tools, and issues motion commands. Its design comprises four progressive layers. To address the modularity challenge, SpaceMind decomposes agent knowledge into independent skill modules with LLM-based dynamic routing, exposes all executable tools through the Model Context Protocol (MCP)~\cite{hou2025mcp} with configurable tool profiles, and switches reasoning modes via injectable mode skills. These three orthogonal dimensions are managed through a single declarative configuration. A Redis message bus decouples the agent from the physical backend, allowing the same codebase to operate across simulation and real hardware without modification. To address the reasoning challenge, SpaceMind hosts three reasoning modes within the same architecture: Standard for direct decision, ReAct~\cite{yao2023react} for iterative thought--action--observation loops, and Prospective for multi-candidate prediction and selection. These modes are complemented by a hierarchical memory system that maintains recent-step detail and long-term compressed summaries. To address the experience accumulation challenge, a Skill Self-Evolution mechanism enables the agent to reflect on each completed episode, distill operational experience into structured learned skill files, and automatically inject them into subsequent runs, thereby accumulating persistent, auditable knowledge without modifying model weights. Finally, to address the validation challenge, we conduct comprehensive closed-loop experiments totaling 192~runs across five satellites, three task types, three reasoning modes, three initial conditions, and two environments, a high-fidelity UE5 simulation and a physical laboratory.

The experiments yield several actionable findings. A tool ablation study of 9~runs shows that removing LiDAR range sensing causes navigation failure, validating the modular tool configuration mechanism. A systematic reasoning evaluation spanning 135~runs under nominal and degraded initial conditions such as underexposure and positional offset reveals that all modes achieve 90--100\% navigation success under nominal conditions but exhibit sharply divergent degradation patterns when rendezvous and search-and-approach are examined separately: Standard maintains 100\% rendezvous success even under degraded conditions but its search success drops to zero. Prospective is the only mode that maintains search success under degradation, while ReAct's rapid termination strategy proves advantageous for inspection under poor visibility. These findings yield concrete mode-selection guidelines for deployment. A self-evolution study covering 30~runs across 6~groups demonstrates that four of six groups achieve significant positive gains. Most notably, one group recovers from complete failure to 100\% success and another raises its inspection score from 12 to 59 out of 100, in both cases by generating learned skills from a single failed episode. Real-world validation over 18~runs confirms that the identical agent codebase, skill definitions, and tool signatures operate a physical mobile robot with zero code modification, achieving 100\% rendezvous success and demonstrating emergent scale adaptation and behavioral consistency across environments.

The contributions of this paper are summarized as follows:
\begin{enumerate}
\item \textbf{Modular agent architecture for on-orbit servicing.} We propose SpaceMind, a VLM-agent framework tailored for autonomous spacecraft operations that decomposes agent knowledge, tools, and reasoning into three independently extensible dimensions, namely skill modules with dynamic routing, MCP-based tools with configurable profiles, and injectable reasoning-mode skills. An environment-agnostic interface layer enables the identical codebase to operate across a high-fidelity space simulation and a physical laboratory without modification.
\item \textbf{Multi-mode reasoning for multi-phase space tasks.} We implement three switchable reasoning modes, Standard, ReAct, and Prospective, within a unified architecture, providing configurable cognitive depth for tasks ranging from straightforward rendezvous to long-horizon search-and-approach under degraded conditions. A hierarchical memory system supports cross-step context reasoning for target-loss recovery and strategy adjustment over extended episodes.
\item \textbf{Skill Self-Evolution for autonomous on-orbit improvement.} We introduce a mechanism that enables the agent to autonomously distill operational experience into persistent, structured skill files after each servicing episode, without model fine-tuning or reward function design, achieving significant recovery from failure in four of six experimental groups.
\item \textbf{Comprehensive closed-loop validation.} We conduct the first systematic evaluation of a VLM-based space agent, spanning 192~runs across five satellites, three tasks, three reasoning modes, and two complementary environments, providing empirical grounding for each architectural design decision.
\end{enumerate}

\section{Related Work}
\label{sec:related}

\subsection{Autonomous Spacecraft Operations and VLM-based Space Agents}
\label{sec:rw_space}

Autonomous on-orbit servicing has traditionally relied on pre-programmed sequences for known cooperative targets, with perception handled by dedicated pipelines for spacecraft detection~\cite{dung2021spacecraft}, part-level semantic segmentation~\cite{zhao2022segmentation,shao2023satellite}, and relative pose estimation~\cite{kisantal2020satellite,park2022speed,park2024spnv2}. These methods have advanced substantially in recent years, as surveyed by Pauly~et~al.~\cite{pauly2023survey}, yet they operate as isolated modules that must be manually integrated into mission-specific control flows, limiting adaptability to new targets or tasks.

A recent line of work has begun exploring large language and vision-language models as high-level spacecraft operators. Rodriguez-Fernandez~et~al.~\cite{rodriguez2024lmspacecraft} demonstrated that language models can decompose spacecraft control tasks through natural-language reasoning, and Carrasco~et~al.~\cite{carrasco2025vlm} extended this to visual-language models for operator-level decision-making. However, these studies share three critical limitations. They address single tasks with monolithic prompts and no modular framework, providing no path to multi-task scalability. They operate in simplified game environments by reading screen text and user-interface elements rather than processing actual visual sensor data, making them text-based agents rather than truly embodied, sensor-driven systems. They also report results on a small number of runs with no cross-environment validation. Separately, Foutter~et~al.~\cite{foutter2024space} explored adapting a foundation model for Mars surface navigation, but without closed-loop control or multi-task evaluation. In contrast, SpaceMind processes raw RGB camera and LiDAR data through a modular architecture, operates in both high-fidelity UE5 simulation and a physical laboratory, and is validated across 192~closed-loop runs spanning five satellite targets and three task types.

\subsection{LLM/VLM-based Embodied Agents and Tool Integration}
\label{sec:rw_agents}

The broader AI agent community has made rapid progress in building autonomous systems around large foundation models. Wang~et~al.~\cite{wang2024llmagent} survey the emerging paradigm of LLM-based agents comprising a language model core, memory, tool use, and planning capabilities. On the tool-integration front, Qin~et~al.~\cite{qin2024toollearning} formalize the framework of tool learning with foundation models, while the Model Context Protocol (MCP)~\cite{hou2025mcp} provides a standardized interface for connecting agents to external tools and data sources. Chain-of-thought prompting~\cite{wei2022chain} showed that explicit intermediate reasoning steps substantially improve LLM decision quality, and Tree of Thoughts~\cite{yao2023tree} extended this idea to deliberate search over multiple reasoning branches. ReAct~\cite{yao2023react} further introduced the influential pattern of interleaving reasoning traces with tool actions, enabling agents to ground their decisions in environmental feedback.

In embodied settings, PaLM-E~\cite{driess2023palme} demonstrated multimodal reasoning grounded in robotic observations, SayCan~\cite{ahn2022saycan} and Inner Monologue~\cite{huang2023inner} enabled language-guided tool use with environmental feedback, and vision-language-action (VLA) models such as RT-2~\cite{brohan2023rt2} and OpenVLA~\cite{kim2024openvla} have demonstrated end-to-end visuomotor control for robotic manipulation. GPT-Driver~\cite{mao2023gptdriver} applied LLM-based planning to autonomous driving. These approaches have primarily targeted terrestrial environments with relatively short task horizons and low-dimensional action spaces (e.g., tabletop grasping, lane following). Three gaps remain when extending this paradigm to space operations. None of these systems addresses 6-DoF long-horizon tasks in unstructured 3D environments. Most adopt a single, fixed reasoning strategy, with no systematic comparison of alternative reasoning modes. Furthermore, tool integration is typically hard-coded for a specific environment, lacking the abstraction layer needed for cross-environment portability. SpaceMind addresses these gaps through an MCP-Redis interface architecture that decouples the agent from the physical backend, a modular skill layer with LLM-based dynamic routing that replaces monolithic prompts, and three switchable reasoning modes evaluated systematically across 135~runs.

\subsection{Experience-driven Self-Improvement for Agents}
\label{sec:rw_evolution}

Enabling agents to improve autonomously from deployment experience is a long-standing goal. Existing adaptation paradigms each carry significant limitations in the context of space operations. Fine-tuning requires labeled training data and gradient computation, both impractical on orbit. Standard in-context learning is bounded by the context window and does not persist across sessions. Retrieval-augmented generation (RAG) retrieves existing documents but cannot synthesize new operational knowledge from experience. Reinforcement learning demands explicit reward function design and is sample-inefficient in physical environments.

Several recent works have explored lighter-weight self-improvement mechanisms for language agents. Reflexion~\cite{shinn2023reflexion} introduced verbal reinforcement learning, where an agent reflects on its failures and stores textual feedback for subsequent trials. However, the reflections remain in the context window and are lost across sessions. ExpeL~\cite{zhao2024expel} extracts reusable ``experiences'' from trajectories and stores them as persistent insights, yet the extracted knowledge remains natural-language summaries without structured format or quality verification. Voyager~\cite{wang2023voyager} builds a persistent skill library by generating executable code functions in Minecraft, representing the closest conceptual precedent to our approach; yet it operates in a digital sandbox with deterministic physics, and its skills are code snippets rather than structured operational knowledge that can be audited or composed with existing domain expertise.

SpaceMind's Skill Self-Evolution mechanism occupies a distinct niche: it generates persistent, structured, auditable operational knowledge from real deployment experience in physical-world space tasks. The learned skills share the same format as hand-authored skills, enabling seamless integration with the modular skill architecture and automatic routing by the skill gateway. Unlike Reflexion, the knowledge persists across sessions; unlike Voyager, it targets physical 3D operations with safety-critical constraints and includes a quality gate with fingerprint deduplication, safety phrase filtering, and task-scope binding to prevent cross-task contamination. Experimental validation across 30~runs in six groups demonstrates that four groups achieve significant improvement, with the mechanism proving particularly effective at recovering from failure and iteratively refining perception strategies.

\section{Methodology}
\label{sec:method}

\subsection{System Overview}
\label{sec:overview}

\begin{figure*}[!t]
  \centering
  \includegraphics[width=\linewidth]{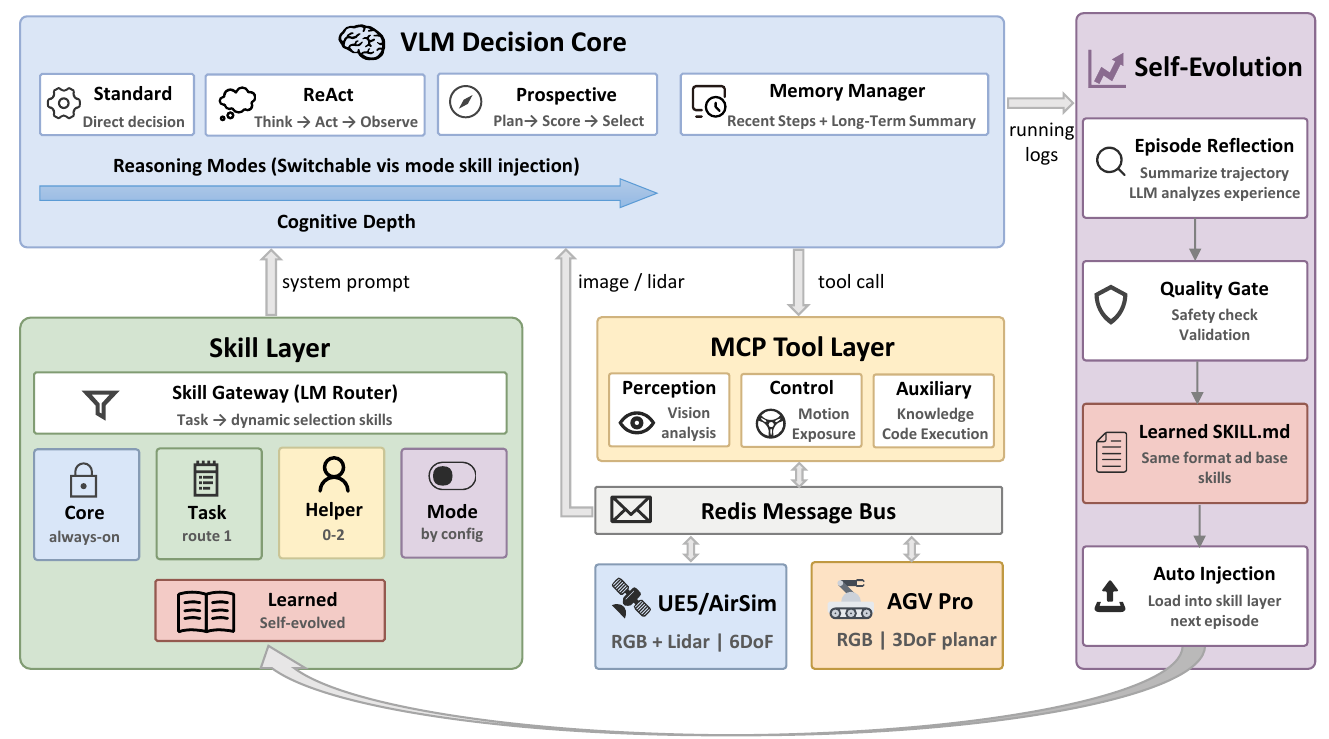}
  \caption{SpaceMind architecture. The framework comprises four layers. The \emph{Skill Layer} decomposes agent knowledge into modular skill definitions with LLM-based dynamic routing~(Sec.~\ref{sec:skill_layer}). The \emph{VLM Decision Core} hosts three switchable reasoning modes with hierarchical memory~(Sec.~\ref{sec:reasoning}). The \emph{MCP Tool Layer} and \emph{Redis Message Bus} provide an environment-agnostic interface that enables the same codebase to operate across UE5 simulation and a physical laboratory~(Sec.~\ref{sec:interface}). A side loop implements \emph{Skill Self-Evolution}, which distills episode experience into learned skill files and injects them into subsequent runs~(Sec.~\ref{sec:tta_method}).}
  \label{fig:architecture}
\end{figure*}

SpaceMind is a closed-loop embodied agent framework in which a vision-language model (VLM) serves as a central decision-control hub. Fig.~\ref{fig:architecture} illustrates the overall architecture. At each decision step~$t$, the agent receives an observation~$o_t$ consisting of an RGB image and an optional LiDAR range summary from the environment. A system prompt is assembled by composing dynamically routed skill modules, reasoning-mode instructions, and a hierarchical memory context summarizing prior steps. The VLM processes this prompt together with the visual observation to produce an action~$a_t$ in the form of a tool call (e.g., \emph{set\_position}, \emph{set\_attitude}, \emph{set\_exposure}), which is dispatched through the Model Context Protocol (MCP)~\cite{hou2025mcp} and relayed to the physical environment via a Redis message bus. The environment executes the action, produces a new observation~$o_{t+1}$, and the loop continues until a termination condition is reached. Formally, the decision process at each step is
\begin{equation}
a_t = \pi_{\text{VLM}}\!\bigl(o_t,\;\mathcal{S}_t,\;\mathcal{M}_t;\;\theta\bigr),
\label{eq:decision}
\end{equation}
where $\mathcal{S}_t$ denotes the assembled skill prompt (determined by the skill gateway's routing result), $\mathcal{M}_t$ the hierarchical memory context, and $\theta$ the frozen VLM parameters. The environment transition follows
\begin{equation}
o_{t+1} = \mathcal{E}\!\bigl(o_t,\;a_t\bigr),
\label{eq:transition}
\end{equation}
where $\mathcal{E}$ abstracts the physical backend (UE5 simulation or laboratory hardware), decoupled from the agent through the Redis message bus.

The architecture is organized around four design layers, each addressing a specific challenge identified in Section~\ref{sec:intro}. The \emph{Skill Layer} with dynamic routing addresses the modularity challenge by decomposing agent knowledge into independently composable skill modules (Sec.~\ref{sec:skill_layer}). The \emph{Reasoning and Memory} layer addresses the cognitive-depth challenge by hosting three switchable reasoning modes and a hierarchical memory system (Sec.~\ref{sec:reasoning}). The \emph{Interface and Communication Architecture} addresses cross-environment portability through MCP tool abstraction and a Redis message bus (Sec.~\ref{sec:interface}). The \emph{Skill Self-Evolution} mechanism addresses experience accumulation by generating persistent learned skills from deployment episodes (Sec.~\ref{sec:tta_method}).

Algorithm~\ref{alg:episode} summarizes the complete execution flow of a single episode, providing a unified view of the four layers before each is detailed in the following subsections.

\begin{algorithm}[htbp]
\caption{SpaceMind Episode Execution}
\label{alg:episode}
\begin{algorithmic}[1]
\Input task description $d$, reasoning mode $m$, tool profile $\mathcal{P}$, skill catalog $\mathcal{C}$, max steps $T$
\Output trajectory $\tau$, outcome $r$
\Statex \textit{// Phase 1: Initialization}
\State $(s_{\mathrm{task}},\,S_{\mathrm{helper}}) \gets \proc{SkillGateway}(d,\,\mathcal{P},\,m,\,\mathcal{C})$
\State $\mathcal{S} \gets S_{\mathrm{core}} \cup \{s_{\mathrm{task}}\} \cup S_{\mathrm{helper}} \cup \proc{ModeSkill}(m) \cup \proc{LearnedSkills}(d)$
\State $\mathcal{M}_0 \gets \varnothing$; \quad $o_0 \gets \mathcal{E}.\proc{Reset}()$
\Statex \textit{// Phase 2: Decision Loop}
\For{$t = 0, 1, \ldots, T\!-\!1$}
    \If{$m = \text{Standard}$}
        \State $a_t \gets \pi_{\text{VLM}}(o_t,\,\mathcal{S},\,\mathcal{M}_t;\,\theta)$ \Comment{Eq.~\eqref{eq:decision}}
    \ElsIf{$m = \text{ReAct}$}
        \State $a_t \gets \proc{ReActLoop}(o_t,\,\mathcal{S},\,\mathcal{M}_t,\,R\!=\!3)$
    \ElsIf{$m = \text{Prospective}$}
        \State $\{c_1,c_2,c_3\} \gets \proc{VLMPlan}(o_t,\,\mathcal{S},\,\mathcal{M}_t)$
        \State $a_t \gets \proc{VLMSelect}(\{c_1,c_2,c_3\},\,o_t)$
    \EndIf
    \State $\textit{result} \gets \proc{MCP.Execute}(a_t)$ \Comment{tool call via Redis}
    \State $o_{t+1} \gets \mathcal{E}.\proc{Observe}()$
    \State $\mathcal{M}_{t+1} \gets \proc{UpdateMemory}(\mathcal{M}_t,\,a_t,\,\textit{result},\,o_{t+1})$
    \If{$\proc{IsTerminated}(a_t,\,\textit{result})$} \textbf{break} \EndIf
\EndFor
\Statex \textit{// Phase 3: Post-Episode Self-Evolution}
\State $\tau \gets \proc{CollectTrajectory}()$; \quad $r \gets \proc{Evaluate}(\tau)$
\State $\textit{mut} \gets \proc{VLMReflect}(\proc{Summarize}(\tau, r),\,\mathcal{S},\,\textit{history})$
\If{$\textit{mut} \neq \proc{NoChange}$ \textbf{and} $\proc{QualityGate}(\textit{mut})$}
    \State $S_{\mathrm{learned}} \gets S_{\mathrm{learned}} \cup \proc{Materialize}(\textit{mut})$
\EndIf
\State \Return $\tau,\; r$
\end{algorithmic}
\end{algorithm}

\subsection{Skill Layer and Dynamic Routing}
\label{sec:skill_layer}

Existing VLM-based agent approaches for space operations encode all task knowledge in a single monolithic system prompt, which creates three problems as the operational scope expands: the prompt becomes unmanageable as more tasks are added, shared strategies (e.g., distance-dependent step sizing, target-loss recovery) cannot be reused across tasks, and experience accumulated during operation has no structured container for storage. To address these limitations, SpaceMind decomposes agent knowledge into independent \emph{skill modules} and routes them dynamically based on the task context.

Each skill is a self-contained knowledge unit consisting of a structured header and a natural-language body. The header declares the skill's name, category, a routing summary describing its applicability, and routing keywords for catalog matching. The body contains the operational instructions that are injected verbatim into the VLM's system prompt. Skills are organized into a three-tier taxonomy:
\begin{itemize}
\item \textbf{Core skills} are always active and provide foundational operational knowledge shared across all tasks, including coordinate conventions, safety constraints, and general action strategies.
\item \textbf{Task skills} encode the primary strategy for a specific task type (e.g., rendezvous, search-and-approach, inspection). Exactly one task skill is selected per episode.
\item \textbf{Helper skills} provide auxiliary strategies that complement the primary task skill, such as distance-dependent step-size decay, target-loss recovery procedures, or multi-modal perception workflows. Zero to two helper skills are selected per episode.
\end{itemize}

A \emph{skill gateway} performs dynamic routing at the start of each episode. The gateway takes the current task description, active tool profile, and reasoning mode as input, and issues a lightweight LLM call over a catalog of available task and helper skills. The LLM selects one primary task skill and up to two complementary helper skills, along with a natural-language justification for its selection. If the routing call fails (e.g., due to malformed output), the gateway falls back to a default skill combination declared in the framework configuration, ensuring robustness.

In addition to task and helper skills, \emph{mode skills} provide reasoning-mode-specific instructions (detailed in Sec.~\ref{sec:reasoning}). These are injected based on the active reasoning mode and are fully decoupled from task skills, so that any combination of task and reasoning mode can be composed without interference. Skills generated by the self-evolution mechanism (Sec.~\ref{sec:tta_method}) share the same format as hand-authored skills, enabling seamless injection through the same routing pipeline.

The final system prompt is assembled by concatenating the active core skills, the routed task and helper skills, the mode-specific skills, and any applicable learned skills, in that order. All skill assignments, tool profiles, reasoning modes, and gateway parameters are managed through a single declarative configuration file, achieving four-dimensional decoupling: any combination of task, reasoning mode, tool set, and skill composition can be specified without code changes.

\subsection{Reasoning Modes and Memory}
\label{sec:reasoning}

Different operational scenarios demand different levels of cognitive depth. A straightforward rendezvous with a visible target may require only direct action selection, whereas navigating under poor visibility or searching for an out-of-view target benefits from deliberative multi-step reasoning. SpaceMind addresses this by hosting three reasoning modes within the same architecture, forming a cognitive-depth gradient that can be selected per episode.

\paragraph{Standard mode} The VLM receives the current observation, the assembled system prompt, and the list of available tools, and directly produces a single tool call as its action. If the selected tool is a perception operation (e.g., image analysis or segmentation), its result can trigger a follow-up VLM call to select a subsequent motion action within the same step. This mode incurs the lowest computational overhead and is suited for tasks where the correct action can be determined from the current observation alone.

\paragraph{ReAct mode} Following the ReAct paradigm~\cite{yao2023react}, each decision step allows up to three internal rounds of \emph{Thought}--\emph{Action}--\emph{Observation}. The agent first generates an explicit reasoning trace, then selects and executes a tool, and observes the result before deciding whether to continue reasoning or commit to a motion action. The inner loop terminates when a motion or termination tool is called. This mode is activated by injecting a mode skill that instructs the VLM to interleave reasoning with tool use, and is suited for tasks requiring information gathering before committing to an action.

\paragraph{Prospective mode} Inspired by deliberative search strategies such as Tree of Thoughts~\cite{yao2023tree}, the agent employs a two-phase deliberation process for high-uncertainty scenarios. In the \emph{planning phase}, the VLM generates three candidate actions, each accompanied by a predicted outcome and a risk assessment. In the \emph{selection phase}, a second VLM call evaluates the candidates and selects the one with the most favorable predicted outcome. This mode is activated by injecting two complementary mode skills, one for planning and one for selection, and provides the highest cognitive depth, enabling the agent to reason about action consequences before committing. If the planning phase produces malformed output, the system gracefully degrades to a direct selection call.

All three modes are activated exclusively through mode skill injection and are fully decoupled from task-specific skills. Switching between modes requires only a configuration change, with no modification to the agent's decision loop or task knowledge.

\paragraph{Hierarchical memory} To support cross-step reasoning, SpaceMind maintains a two-tier memory system. The \emph{recent memory} retains detailed records of the most recent $N$ steps, including the agent's analysis, tool selections, arguments, and results. The \emph{long-term memory} compresses older steps into a natural-language summary that captures cumulative trajectory information. Both tiers are concatenated and injected as contextual input to each VLM call, enabling the agent to detect and recover from target loss, avoid repeating failed strategies, and track progress toward task completion across extended episodes.

\subsection{Interface and Communication Architecture}
\label{sec:interface}

A key design goal of SpaceMind is that the same agent codebase operates across different physical environments without modification. Achieving this requires decoupling the agent's decision logic from the specifics of any particular sensor or actuator backend. SpaceMind implements this decoupling through two abstraction layers: an agent-tool protocol and a tool-environment communication bus.

\paragraph{MCP tool protocol} All executable capabilities are exposed to the agent as tools through the Model Context Protocol (MCP)~\cite{hou2025mcp}, a standardized interface for connecting language models to external functions. The agent invokes tools by name and argument via a unified calling convention, without knowledge of how each tool is implemented. Tools are organized into four functional categories:
\begin{itemize}
\item \emph{Perception tools} analyze the current visual input, including brightness assessment, part-level segmentation, region cropping, and zoom.
\item \emph{Control tools} issue motion and sensor commands such as translational and rotational motion, camera exposure adjustment, and task termination.
\item \emph{Knowledge tools} provide access to a domain knowledge base describing spacecraft characteristics.
\item \emph{Auxiliary tools} support optional capabilities such as code execution for numerical computation.
\end{itemize}
A \emph{tool profile} mechanism controls which tools are visible to the agent for a given experiment. Each profile specifies an allowed-tool whitelist; at runtime, only tools appearing in both the MCP server's full catalog and the active profile's whitelist are presented to the VLM. This enables controlled ablation studies (e.g., removing range sensing to test vision-only operation) and environment-specific tool sets (e.g., a minimal profile for the physical laboratory) through configuration alone.

Two design decisions merit discussion. First, translational and rotational motion are exposed as separate tools rather than a single 6-DoF command, preventing the VLM from producing coupled translation-rotation actions that lead to unintended spiral trajectories, a failure mode observed in early development. Second, camera exposure adjustment is exposed as an explicit tool, enabling the agent to actively improve its own perception quality when it detects underexposure or overexposure, rather than passively accepting degraded imagery.

\paragraph{Redis message bus} Between the MCP tool implementations and the physical environment, a Redis publish-subscribe message bus provides the second abstraction layer. A unified naming contract defines the communication channels: sensor data flows upstream through designated keys (e.g., latest RGB image, latest LiDAR summary) and image-arrival notifications, while control commands flow downstream through corresponding command topics (e.g., pose-change increments, exposure adjustments). In the UE5 simulation, an AirSim~\cite{shah2018airsim} bridge process writes sensor data to Redis and consumes motion commands. In the physical laboratory, a ROS2 sensor bridge publishes camera frames to the same Redis keys, and a separate motion executor subscribes to the same command topics to drive the mobile robot. Both backends populate and consume identical Redis channels; the agent's decision loop contains no environment-conditional logic.

This two-layer abstraction yields a concrete portability guarantee: the same agent decision loop, MCP tool server, and communication contract operate in both environments with zero code modification. Extending SpaceMind to a new environment backend requires only implementing the Redis read/write endpoints for that backend's sensors and actuators.

\subsection{Skill Self-Evolution Mechanism}
\label{sec:tta_method}

Space on-orbit servicing is inherently a repeated-execution domain: the same task types recur across different targets and conditions. An agent that discards all operational experience after each episode forfeits the opportunity to improve from its own successes and failures. Traditional adaptation methods are poorly suited to this setting: model fine-tuning requires labeled training data and gradient computation, reinforcement learning demands explicit reward function design, in-context learning is bounded by the context window and does not persist across sessions, and retrieval-augmented generation can only retrieve existing documents rather than synthesize new knowledge. SpaceMind introduces a \emph{Skill Self-Evolution} mechanism that enables the agent to autonomously generate, validate, and accumulate operational knowledge across episodes without modifying model weights.

The mechanism operates as an outer loop around the standard decision cycle (Fig.~\ref{fig:architecture}, right side). After each episode concludes, the following pipeline executes:

\paragraph{Episode summarization} The system records the complete tool-call trajectory, success or failure outcome, termination reason, and a structured summary of the agent's movement and perception history during the episode.

\paragraph{Experience reflection} The episode summary, together with summaries from recent prior episodes and the currently active skill definitions, is provided to the VLM in a dedicated reflection call. The VLM analyzes what worked, what failed, and why, and produces a structured \emph{mutation decision} specifying one of five actions: \emph{create} a new learned skill capturing a previously unknown operational pattern; \emph{overlay} an existing skill with refined or supplementary rules; \emph{rewrite} an existing skill when evidence warrants a fundamental revision; \emph{disable} a learned skill that has been found harmful; or \emph{no\_change} when the current experience does not justify a modification.

\paragraph{Quality gate} Every proposed mutation passes through a multi-stage validation gate before being applied. The gate enforces four constraints:
\begin{itemize}
\item A safety-phrase blacklist rejects mutations containing instructions that could override safety rules.
\item A fingerprint deduplication check prevents semantically redundant skills from accumulating.
\item A task-scope binding ensures that skills learned in one task context cannot contaminate another task's knowledge base.
\item A parent-skill validation confirms that overlay and rewrite operations target skills that actually exist.
\end{itemize}
Mutations that fail any constraint are discarded, and an audit log records the rejection reason.

\paragraph{Learned skill generation} Mutations that pass the quality gate are materialized as new skill files sharing the same format as hand-authored skills, ensuring compatibility with the existing skill routing infrastructure. Each learned skill contains five structured sections: \emph{Intent} specifies what the skill aims to achieve, \emph{Trigger} defines the conditions under which the skill should activate, \emph{Rule} encodes the operational instructions, \emph{Constraints} delineates safety boundaries, and \emph{Evidence} records the episode observations that motivated the skill. Additional metadata records the skill's provenance, version, originating episode, and applicable task scope.

\paragraph{Automatic injection} At the start of each subsequent episode, the skill runtime loads all learned skills alongside the base skill set. A retrieval function matches learned skills to the current task and reasoning mode based on their declared scope and trigger conditions, selecting the top-$k$ most relevant learned skills for injection into the system prompt alongside the dynamically routed base skills.

This design has four properties that distinguish it from prior self-improvement approaches.
First, learned skills are \emph{naturally compatible} with the modular skill architecture (Sec.~\ref{sec:skill_layer}): they use the same data format, pass through the same routing pipeline, and compose with base skills without special handling.
Second, the knowledge is \emph{persistent and auditable}: each mutation is recorded with full provenance, and the skill files can be inspected, edited, or reverted by human operators.
Third, the evolution is \emph{progressive}: skills accumulate incrementally through versioned overlays rather than wholesale replacement, allowing gradual refinement.
Fourth, the quality gate provides a \emph{safety guarantee}: even if the VLM's reflection produces an unsafe or redundant suggestion, the hard constraints prevent it from entering the active skill set.

\section{Experimental Validation}
\label{sec:experiments}

We evaluate SpaceMind through 192 closed-loop runs organized into four experimental campaigns, conducted across a high-fidelity Unreal Engine~5 (UE5) simulation and a physical laboratory environment. Table~\ref{tab:overview} summarizes the campaigns. The tool ablation campaign validates the modular tool configuration mechanism. The reasoning evaluation campaign provides a systematic comparison of reasoning modes across five satellites, three tasks, and three initial conditions. The self-evolution campaign tests autonomous skill improvement through iterative execution. The real-world campaign validates cross-environment portability with zero code modification.

\begin{table}[htbp]
  \centering
  \caption{Overview of experimental campaigns.}
  \label{tab:overview}
  \small
  \begin{tabular}{llc}
    \toprule
    \textbf{Campaign} & \textbf{Env} & \textbf{Runs} \\
    \midrule
    Tool Ablation     & UE5     & 9   \\
    Reasoning Eval    & UE5     & 135 \\
    Self-Evolution    & UE5     & 30  \\
    Real-World        & AGV Pro & 18  \\
    \midrule
    \textbf{Total}    &         & \textbf{192} \\
    \bottomrule
  \end{tabular}
\end{table}

\subsection{Experimental Setup}
\label{sec:setup}

\paragraph{Simulation environment}
The simulation is built on UE5 with the AirSim plugin~\cite{shah2018airsim}, extending the SpaceSense-Bench platform~\cite{wu2026spacesensebench} to provide high-fidelity optical rendering and multi-sensor simulation. Five satellite models of varying geometry and scale are placed in the scene: CAPSTONE, IBEX, BioSentinel, New Horizons, and Huygens. The agent perceives the environment through an RGB camera and a LiDAR range sensor, and controls its 6-DoF position and attitude via MCP~\cite{hou2025mcp} tool calls relayed over a Redis message bus.

\paragraph{Laboratory environment}
The physical testbed consists of a myAGV Pro omnidirectional mobile robot with Mecanum wheels and a 40\,kg payload, equipped with an Orbbec Gemini~2 stereo RGB camera at $1920 \times 1080$ resolution. The robot runs ROS2 Humble and communicates with the same agent decision loop through a Redis bridge. Two 1-meter-diameter 3D-printed satellite mockups, CAPSTONE and Artemis, serve as targets. Motion is constrained to a 3-DoF planar workspace.

\paragraph{Tasks}
Three task types of increasing difficulty are defined in Table~\ref{tab:tasks}.

\begin{table}[htbp]
  \centering
  \caption{Task definitions and evaluation metrics.}
  \label{tab:tasks}
  \small
  \resizebox{\linewidth}{!}{%
  \begin{tabular}{lcll}
    \toprule
    \textbf{Task} & \textbf{Abbr.} & \textbf{Description} & \textbf{Metrics} \\
    \midrule
    Rendezvous           & Rndz.  & Target visible; approach to ${\sim}$2\,m   & Succ, Dist, Steps \\
    Search-and-Approach  & Search & Target not visible; find then approach to ${\sim}$2\,m & Succ, Dist, Steps \\
    Inspection           & Insp.  & Close-range 5-dim structured report & Score (0--100) \\
    \bottomrule
  \end{tabular}}
\end{table}

\paragraph{Initial conditions}
Three initial conditions are used in the UE5 simulation to evaluate robustness (Table~\ref{tab:conditions}). All three tasks are tested under each condition.

\begin{table}[htbp]
  \centering
  \caption{Initial condition definitions for UE5 simulation.}
  \label{tab:conditions}
  \small
  \begin{tabular}{clll}
    \toprule
    \textbf{Cond.} & \textbf{Position} & \textbf{Exposure} & \textbf{Difficulty} \\
    \midrule
    C1 & Nominal          & Nominal       & Baseline  \\
    C2 & Far lateral offset & Underexposure & Degraded  \\
    C3 & Reverse offset   & Overexposure  & Degraded  \\
    \bottomrule
  \end{tabular}
\end{table}

\paragraph{VLM backbone}
All experiments use the open-source Qwen3-VL-235B-A22B model~\cite{bai2025qwen3vl} as the VLM decision core, with temperature $\leq 0.3$ to promote output stability across all campaigns.

\subsection{Tool Configuration Ablation}
\label{sec:tool_ablation}

To validate the modular tool configuration mechanism, the tool ablation campaign compares three tool profiles under fixed conditions, specifically Standard mode on CAPSTONE with C1. Each profile exposes a different subset of MCP tools to the agent: \emph{Vision-Only} provides camera perception only, \emph{Hybrid-Nav} adds LiDAR range sensing, and \emph{Hybrid-Nav-Code} further adds runtime code execution. Table~\ref{tab:tool_ablation} presents the results.

\begin{table}[htbp]
  \centering
  \caption{Tool profile ablation results over 9~runs.}
  \label{tab:tool_ablation}
  \small
  \resizebox{\linewidth}{!}{%
  \begin{tabular}{l cc cc c c}
    \toprule
    & \multicolumn{2}{c}{\textbf{Rndz.}} & \multicolumn{2}{c}{\textbf{Search}} & \textbf{Nav.\ Pass} & \textbf{Insp.} \\
    \cmidrule(lr){2-3} \cmidrule(lr){4-5} \cmidrule(lr){6-6} \cmidrule(lr){7-7}
    \textbf{Tool Profile} & Dist\,(m) & Steps & Dist\,(m) & Steps & Rate & Score \\
    \midrule
    Vision-Only       & \multicolumn{2}{c}{FAIL} & \multicolumn{2}{c}{FAIL} & 0/2 & 65 \\
    Hybrid-Nav        & 2.006 & 9  & 2.277 & 16 & \textbf{2/2} & \textbf{65} \\
    Hybrid-Nav-Code   & 2.048 & 13 & 2.279 & 15 & 2/2 & 50 \\
    \bottomrule
  \end{tabular}}
\end{table}

Without LiDAR range measurements, the Vision-Only profile fails both navigation tasks due to distance estimation errors that cause overshoot. Hybrid-Nav succeeds on all tasks with the fewest steps and highest inspection score, while adding runtime code execution does not improve performance, as the agent never invokes the code-execution tool in practice. Based on these results, Hybrid-Nav is adopted as the default for all subsequent experiments, demonstrating that SpaceMind's tool profile mechanism allows declarative capability configuration.

\subsection{Multi-Satellite Reasoning Mode Evaluation}
\label{sec:reasoning_eval}

The reasoning evaluation campaign provides a systematic comparison across 5~satellites, 3~tasks, 3~reasoning modes (Standard, ReAct~\cite{yao2023react}, Prospective), and 3~initial conditions, totaling 135~runs.

\subsubsection{Results}
\label{sec:reasoning_results}

\begin{figure}[htbp]
  \centering
  \includegraphics[width=0.65\linewidth]{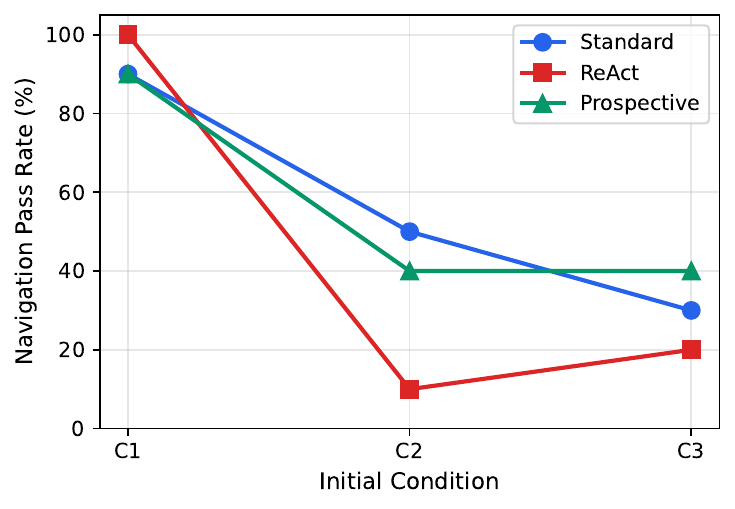}
  \caption{Navigation pass rate across initial conditions. ReAct degrades sharply from C1 to C2, while Standard and Prospective maintain higher robustness.}
  \label{fig:degradation}
\end{figure}

Table~\ref{tab:reasoning_main} consolidates navigation and inspection results across all modes and conditions. Fig.~\ref{fig:degradation} visualizes the navigation pass rate degradation from C1 to C3.

\begin{table}[htbp]
  \centering
  \caption{Reasoning mode evaluation results across conditions, 135~runs in total. Navigation reports pass count per 5~runs for each task type; inspection reports average score per 5~runs. Best values in each column are \textbf{bold}.}
  \label{tab:reasoning_main}
  \small
  \resizebox{\linewidth}{!}{%
  \begin{tabular}{l ccc c ccc c cc ccc c}
    \toprule
    & \multicolumn{4}{c}{\textbf{Rndz.} (pass / 5)} & \multicolumn{4}{c}{\textbf{Search} (pass / 5)} & \multicolumn{2}{c}{\textbf{Nav.\ Total}} & \multicolumn{4}{c}{\textbf{Insp.} (avg score)} \\
    \cmidrule(lr){2-5} \cmidrule(lr){6-9} \cmidrule(lr){10-11} \cmidrule(lr){12-15}
    \textbf{Mode} & C1 & C2 & C3 & Sub & C1 & C2 & C3 & Sub & Pass & Dist\,(m) & C1 & C2 & C3 & Avg \\
    \midrule
    Standard    & 5 & \textbf{5} & \textbf{3} & \textbf{13 (87\%)} & 4 & 0 & 0 & 4 (27\%) & 17/30 (57\%) & 2.143 & \textbf{40.2} & \textbf{41.0} & 36.2 & 39.1 \\
    ReAct       & 5 & 1 & 2 & 8 (53\%) & \textbf{5} & 0 & 0 & 5 (33\%) & 13/30 (43\%) & 2.177 & 38.6 & 40.8 & \textbf{42.8} & \textbf{40.7} \\
    Prospective & 5 & 3 & 2 & 10 (67\%) & 4 & \textbf{1} & \textbf{2} & \textbf{7 (47\%)} & 17/30 (57\%) & \textbf{2.079} & 36.6 & 22.6 & 29.8 & 29.7 \\
    \bottomrule
  \end{tabular}}
\end{table}

\subsubsection{Analysis}
\label{sec:reasoning_analysis}

The results reveal that \emph{no single reasoning mode dominates}, and the Rndz./Search split exposes fundamentally different degradation patterns. All three modes achieve 5/5 rendezvous success under C1 where the target is visible and conditions are nominal. Standard maintains perfect rendezvous even under C2 at 5/5, achieving the highest rendezvous subtotal of 13/15 at 87\%. However, its search performance collapses entirely under degradation: 0/5 at both C2 and C3, yielding only 4/15 or 27\%. ReAct shows a similar pattern, achieving perfect 5/5 rendezvous and search at C1 but dropping to 1/5 rendezvous and 0/5 search under C2, with most failures reaching the 50-step timeout as the Thought--Action--Observation loop oscillates when the target is far from the expected position.

Prospective is the \emph{only} mode that achieves any search success under degraded conditions: 1/5 at C2 and 2/5 at C3, yielding a search subtotal of 7/15 at 47\%, nearly double Standard's 27\%. Its candidate-prediction mechanism avoids blind exploration by evaluating multiple movement hypotheses before committing, yielding the most precise terminal distance of 2.079\,m. This advantage is specific to search-and-approach tasks where the target must first be found; for rendezvous where the target is already visible, Standard suffices and additional reasoning overhead provides no benefit.

For inspection, a counter-intuitive pattern emerges: ReAct achieves the highest score under C3 at 42.8, surpassing Standard at 36.2 and Prospective at 29.8. ReAct's rapid termination strategy of 1--3 steps limits hallucination accumulation, whereas Prospective's two-stage reasoning amplifies VLM hallucination risk under degraded visual conditions. These findings validate that SpaceMind's modular reasoning architecture can host fundamentally different strategies, and the systematic comparison provides actionable mode-selection guidance for deployment.

\subsubsection{Skill Routing Validation}
\label{sec:skill_routing}

To validate the dynamic skill routing mechanism, we extract the routing decisions from all 135~runs. The skill gateway correctly routes each task to its intended skill set: \emph{rendezvous} to approach + distance skills; \emph{search-and-approach} to search + target-recovery + distance skills; and \emph{inspection} to inspection + perception skills. The routing achieves 100\% agreement with the intended assignments across all runs without manual intervention, confirming that the LLM-based skill gateway can reliably compose task-appropriate skill sets from the modular catalog.

\subsection{Skill Self-Evolution}
\label{sec:self_evolution}

\begin{figure}[htbp]
  \centering
  \includegraphics[width=\linewidth]{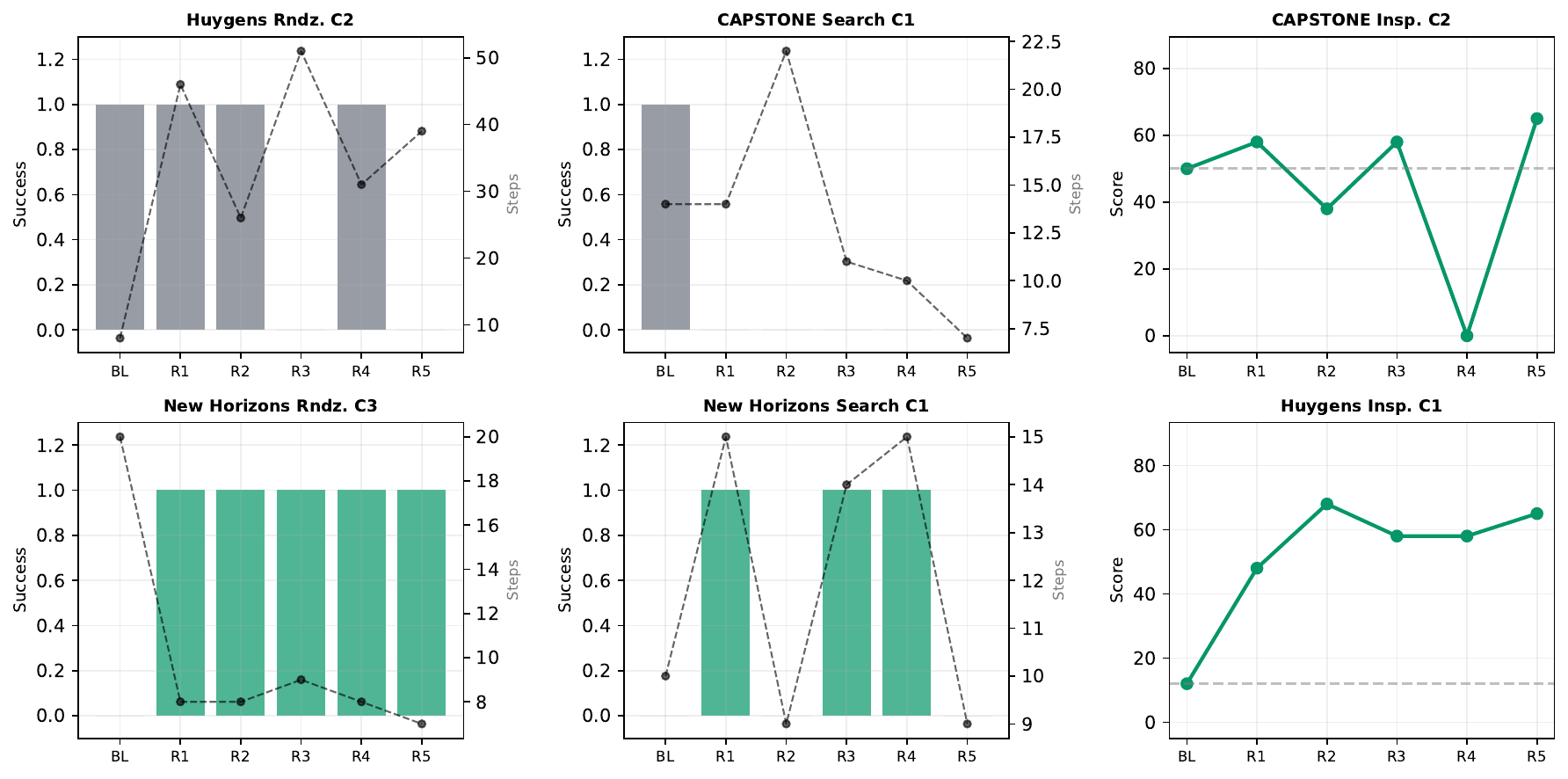}
  \caption{Skill self-evolution learning curves for six groups, organized by task type in columns for Rndz., Search, and Insp.\ respectively. For navigation groups, bars show success and the dashed line shows step count; for inspection groups, the solid line shows score and the dashed horizontal line marks the baseline level. Green indicates clear positive benefit; gray indicates no significant improvement.}
  \label{fig:learning}
\end{figure}

The self-evolution campaign evaluates whether the skill self-evolution mechanism can accumulate useful operational knowledge through repeated execution. We select six groups spanning three satellites and two task types, prioritizing combinations where the reasoning evaluation baseline failed or scored low. Each group undergoes five consecutive rounds with a shared evolution workspace, using Standard mode and the Hybrid-Nav tool profile. After each round, the evolution runtime reflects on the episode and may generate learned skill files that are automatically injected into subsequent rounds. Table~\ref{tab:self_evol} summarizes the results and Fig.~\ref{fig:learning} shows the per-round learning curves.

\begin{table}[htbp]
  \centering
  \caption{Skill self-evolution results over 30~runs across 6 groups. BL = baseline from the reasoning evaluation campaign.}
  \label{tab:self_evol}
  \small
  \resizebox{\linewidth}{!}{%
  \begin{tabular}{l l c rr rr l}
    \toprule
    & & & \multicolumn{2}{c}{\textbf{Baseline}} & \multicolumn{2}{c}{\textbf{Best After Evol.}} & \\
    \cmidrule(lr){4-5} \cmidrule(lr){6-7}
    \textbf{Satellite} & \textbf{Task} & \textbf{Cond} & Result & Detail & Result & Detail & \textbf{Verdict} \\
    \midrule
    CAPSTONE     & Search & C1 & OK    & 2.597\,m & 0/5   & --         & No benefit \\
    CAPSTONE     & Insp.  & C2 & 50\,pt & --       & 65\,pt & +30\%     & Positive \\
    New Horizons & Search & C1 & FAIL  & 10\,st   & 3/5   & 2.102\,m   & Positive \\
    New Horizons & Rndz.  & C3 & FAIL  & 20\,st   & \textbf{5/5} & 7\,st & Strong + \\
    Huygens      & Rndz.  & C2 & OK    & 2.209\,m & 3/5   & 2.169\,m   & No benefit \\
    Huygens      & Insp.  & C1 & 12\,pt & --      & \textbf{68\,pt} & +467\% & Strong + \\
    \bottomrule
  \end{tabular}}
\end{table}

The most striking result is the mechanism's ability to recover from failure. The two groups that fail completely at baseline, New Horizons Search~C1 and New Horizons Rndz.~C3, achieve 60\% and 100\% success rates respectively after self-evolution. In New Horizons Rndz.~C3, the step count converges steadily from 20 to 7 across five rounds, indicating that the automatically generated learned skill encodes an effective distance-dependent step-size reduction strategy. Three of the six groups show significant improvement by Round~1 or~2, demonstrating that the reflection mechanism can extract actionable knowledge from a single failed episode.

For inspection tasks, both groups show consistent gains through the overlay mutation mechanism: CAPSTONE Insp.~C2 improves from 50 to 65~points and Huygens Insp.~C1 from 12 to an average of 59.4~points. The incremental overlay refinement of perception strategies proves particularly effective for this task type.

However, the two groups where the baseline already succeeds, CAPSTONE Search~C1 and Huygens Rndz.~C2, do not benefit from self-evolution, revealing that the mechanism's primary value lies in error correction and strategy discovery rather than optimizing already-successful behavior. Across all groups, seven learned skills are generated in the standard skill-file format and correctly routed by the skill gateway in subsequent rounds, confirming compatibility between the self-evolution mechanism and the modular architecture.

\subsection{Real-World Validation}
\label{sec:realworld}

The real-world campaign validates that the same architecture, including the identical agent loop, MCP tool signatures, and Redis communication contracts, can operate a physical robot with zero code modification; only the physical backend is swapped to a ROS2 sensor bridge and motor controller. Table~\ref{tab:lab} presents the results across two satellite mockups, three tasks, and three repetitions per combination, totaling 18~runs.

\begin{table}[htbp]
  \centering
  \caption{Real-world validation results over 18~runs.}
  \label{tab:lab}
  \small
  \resizebox{\linewidth}{!}{%
  \begin{tabular}{l ccc ccc cc}
    \toprule
    & \multicolumn{3}{c}{\textbf{Rndz.}} & \multicolumn{3}{c}{\textbf{Search}} & \multicolumn{2}{c}{\textbf{Insp.}} \\
    \cmidrule(lr){2-4} \cmidrule(lr){5-7} \cmidrule(lr){8-9}
    \textbf{Satellite} & Pass & Runs & Rate & Pass & Runs & Rate & Avg Score & Runs \\
    \midrule
    CAPSTONE & 3 & 3 & 100\% & 2 & 3 & 67\% & 41.0 & 3 \\
    Artemis  & 3 & 3 & 100\% & 2 & 3 & 67\% & 34.0 & 3 \\
    \midrule
    \textbf{Total} & \textbf{6} & \textbf{6} & \textbf{100\%} & \textbf{4} & \textbf{6} & \textbf{67\%} & \textbf{36.8} & \textbf{6} \\
    \bottomrule
  \end{tabular}}
\end{table}

\begin{figure}[!htb]
  \centering
  \includegraphics[width=\linewidth]{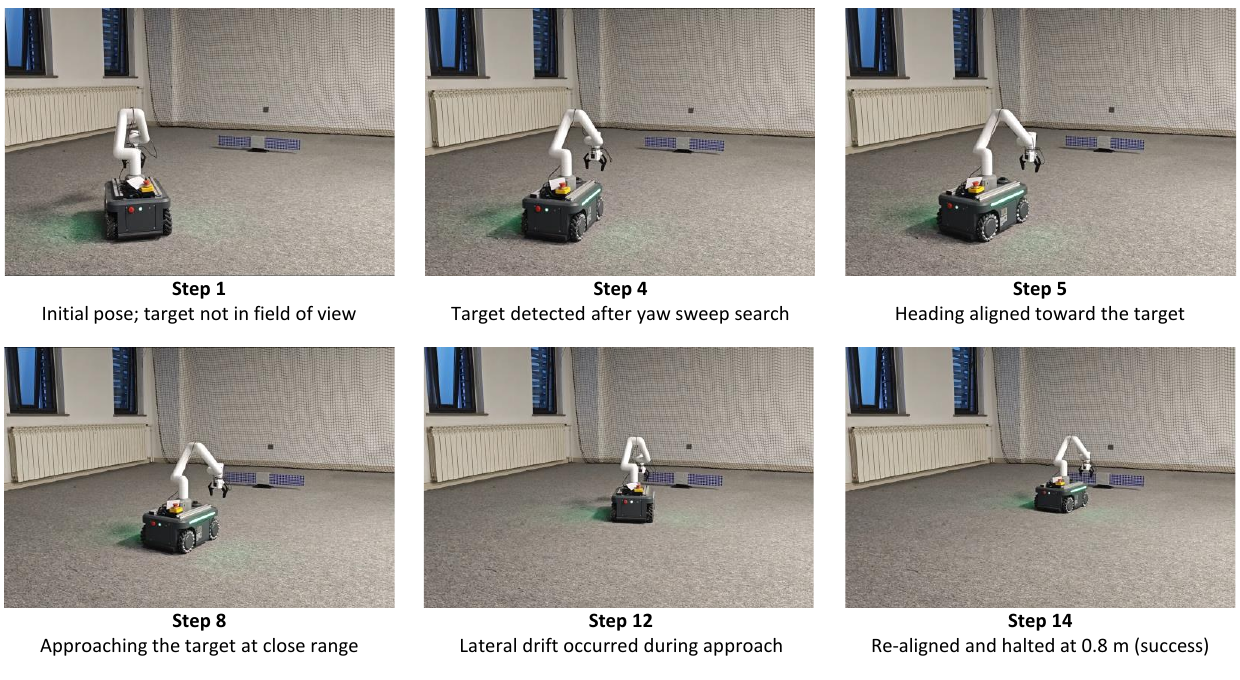}
  \caption{Third-person view of a real-world search-and-approach episode on CAPSTONE over 14~steps. The agent searches for the target via yaw sweep in Steps~1--4, aligns and approaches in Steps~5--8, recovers from lateral drift at Step~12, and terminates at 0.8\,m from the target surface at Step~14.}
  \label{fig:lab_thirdperson}
\end{figure}

The agent achieves a perfect 6/6 pass rate on rendezvous and 4/6 on search-and-approach tasks. Three observations are specific to the physical environment. First, the agent autonomously adapts its step size from 1--2\,m per step in simulation to 0.1--0.2\,m per step in the laboratory, demonstrating emergent scale awareness without explicit re-configuration. Second, the target-recovery skill's yaw $\pm 90^{\circ}$ sweep pattern transfers identically from simulation, confirming behavioral consistency. Third, one inspection run results in a collision due to cumulative small steps over 15 iterations, highlighting the need for explicit safety margins in close-proximity operations. Fig.~\ref{fig:lab_thirdperson} illustrates a representative search-and-approach episode from the third-person perspective. These results validate the interface and communication architecture and demonstrate that SpaceMind's modular design generalizes across environments.

\subsection{Qualitative Analysis}
\label{sec:qualitative}

\begin{figure}[!htb]
  \centering
  \begin{tabular}{@{}c@{\hspace{3pt}}c@{\hspace{3pt}}c@{\hspace{3pt}}c@{}}
    \includegraphics[width=0.235\linewidth]{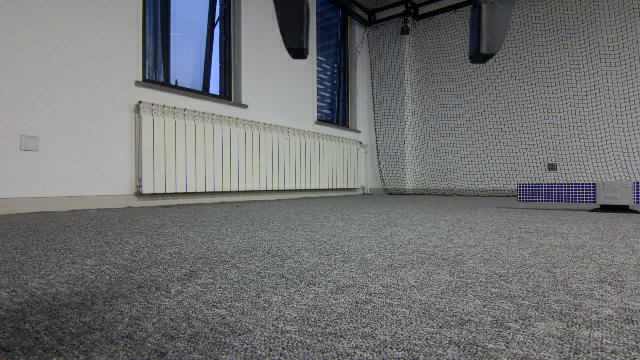} &
    \includegraphics[width=0.235\linewidth]{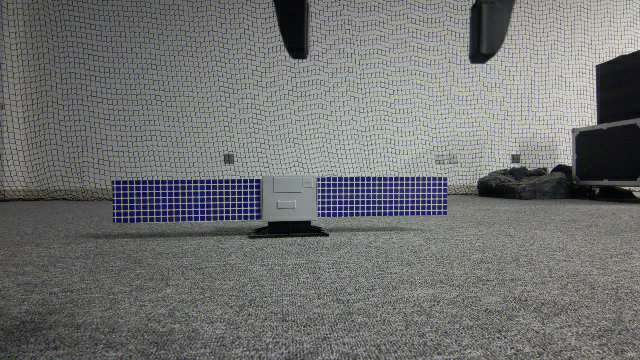} &
    \includegraphics[width=0.235\linewidth]{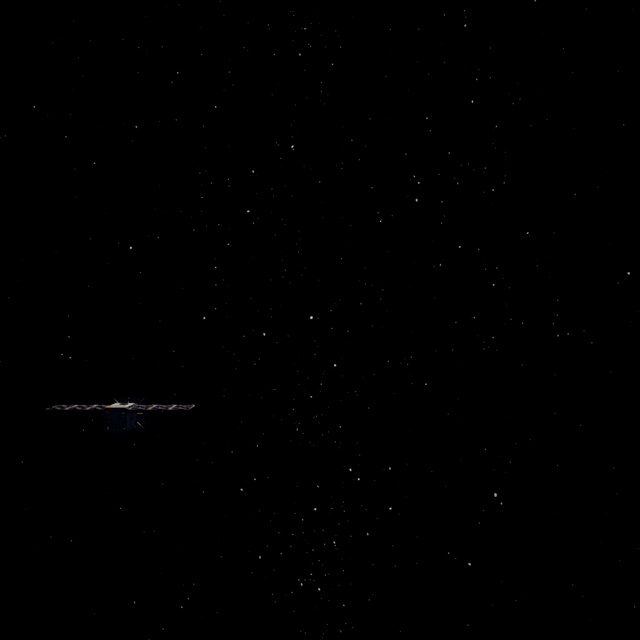} &
    \includegraphics[width=0.235\linewidth]{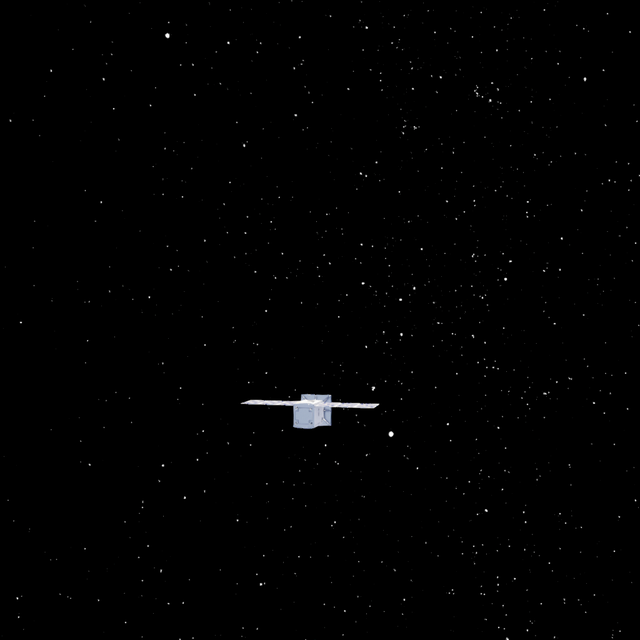} \\
    {\scriptsize (a) Lab: Step 1} & {\scriptsize (b) Lab: Step 42} &
    {\scriptsize (c) Exp: Step 1} & {\scriptsize (d) Exp: Step 10} \\[4pt]
    \includegraphics[width=0.235\linewidth]{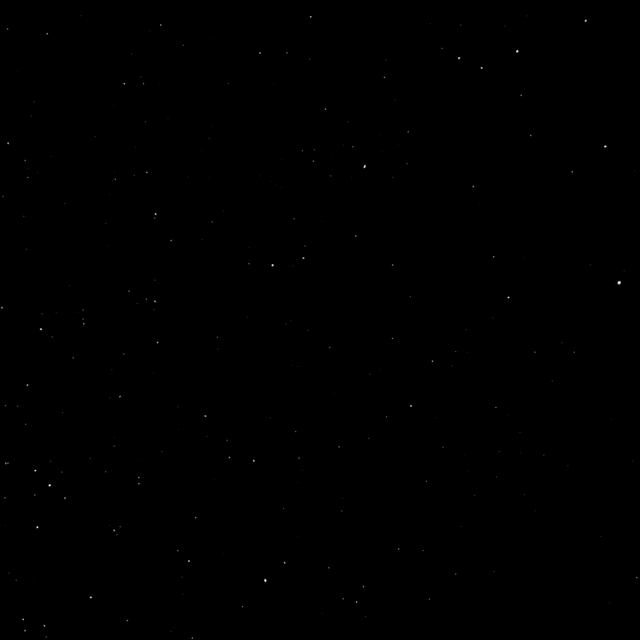} &
    \includegraphics[width=0.235\linewidth]{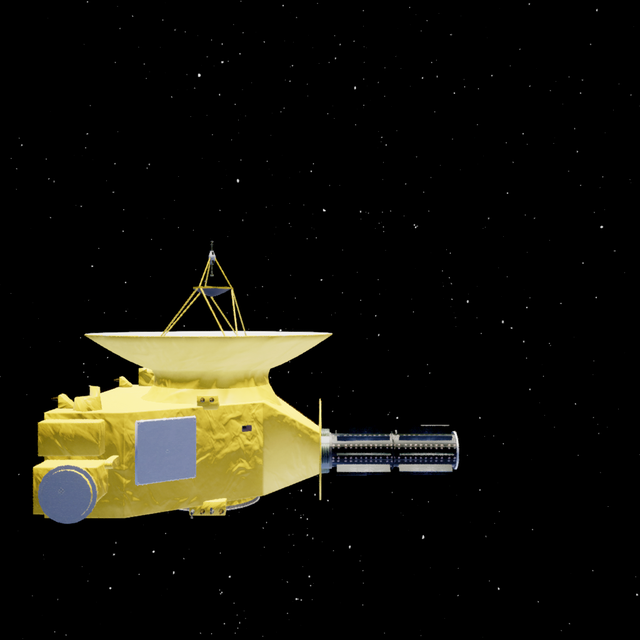} &
    \includegraphics[width=0.235\linewidth]{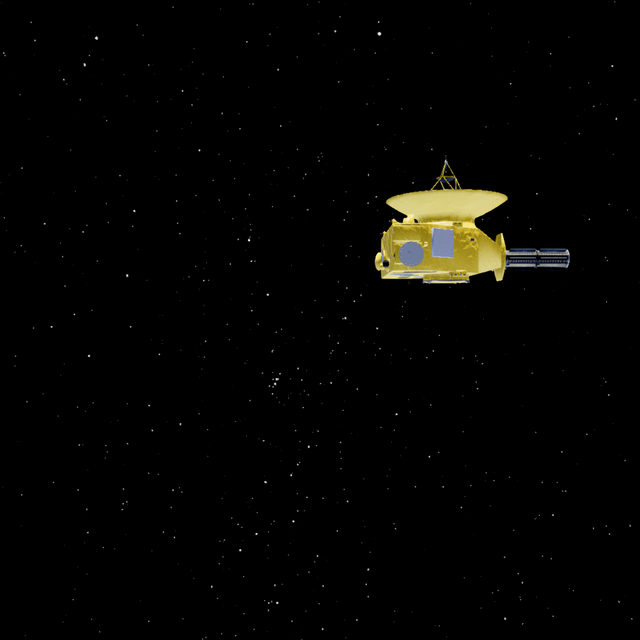} &
    \includegraphics[width=0.235\linewidth]{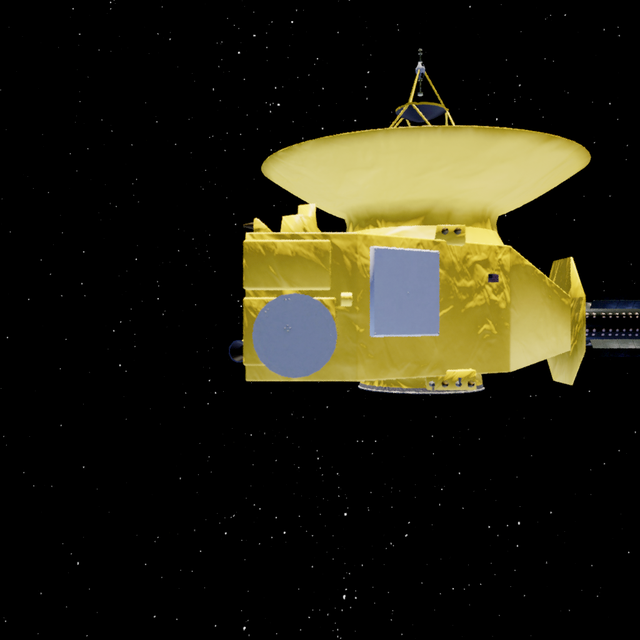} \\
    {\scriptsize (e) Prosp: Step 1} & {\scriptsize (f) Prosp: Step 29} &
    {\scriptsize (g) Evol: Step 1} & {\scriptsize (h) Evol: Step 8} \\
  \end{tabular}
  \caption{Qualitative examples from the agent's sensor view. Top row: (a--b)~Lab search, where the target is not visible at Step~1 but found and approached by Step~42; (c--d)~inspection under C2, where the image is nearly black at Step~1 and clear after active exposure adjustment at Step~10. Bottom row: (e--f)~Prospective search under C2, with an empty view at Step~1 and the target reached at Step~29; (g--h)~self-evolution for New Horizons Rndz.~C3, showing the baseline at Step~1 and successful approach at Step~8 after learned skill injection.}
  \label{fig:qualitative}
\end{figure}

We select four representative episodes that illustrate emergent behaviors enabled by SpaceMind's design. Fig.~\ref{fig:qualitative} shows keyframe images from each episode.

\paragraph{Example 1: Lab target-lost recovery}
During a CAPSTONE search task in the laboratory, the target is not visible from the initial pose (Fig.~\ref{fig:qualitative}a). Guided by the target-recovery skill, the agent executes a bounded yaw sweep of six $15^{\circ}$ right turns followed by nine $15^{\circ}$ left turns before detecting the target and switching to approach mode. The episode spans 42~steps, terminating at 1.13\,m, demonstrating the interplay between modular skill knowledge and the hierarchical memory system.

\paragraph{Example 2: Active exposure adjustment}
Under the underexposed C2 condition, the initial camera image is nearly black. The agent invokes a brightness diagnostic tool, then calls the exposure-control tool to increase the camera exposure. After adjustment, the target becomes clearly visible (Fig.~\ref{fig:qualitative}c--d), and the agent proceeds with inspection. This highlights SpaceMind's role as a decision-control hub that actively adjusts sensor parameters rather than passively consuming degraded imagery.

\paragraph{Example 3: Prospective candidate prediction}
In a New Horizons search task under C2, both Standard and ReAct fail with 50-step timeouts, but Prospective succeeds in 29~steps at 2.007\,m (Fig.~\ref{fig:qualitative}e--f). The planner generates three candidate movement directions with predicted outcomes and risk scores; the selector chooses the lowest-risk option, avoiding the blind exploration that causes other modes to time out.

\paragraph{Example 4: Self-evolution failure recovery}
In the New Horizons Rndz.~C3 group, the baseline fails at 20~steps. After one round of self-evolution, the evolution runtime generates a learned skill encoding a distance-dependent step-size reduction strategy. The agent succeeds consistently from Round~1 onward, with step counts converging from 20 to 7 across five rounds (Fig.~\ref{fig:qualitative}h). The learned skill is persisted as a standard skill file, making it auditable, portable, and automatically routed by the skill gateway.

\section{Discussion}
\label{sec:discussion}

\paragraph{Framework positioning and modularity}
SpaceMind is designed as a modular, self-evolving embodied agent \emph{infrastructure} for space operations. Its value lies in progressive layering: modular knowledge, tools, and reasoning; multi-mode reasoning with hierarchical memory; skill self-evolution; and closed-loop cross-environment validation. The experimental results confirm this design: the skill routing gateway achieves 100\% correct task-to-skill composition across all 135~runs; the tool ablation study validates plug-and-play capability configuration; and the real-world campaign demonstrates zero-code-modification cross-environment transfer. Notably, the agent exhibits emergent scale awareness, autonomously reducing its step size from 1--2\,m in simulation to 0.1--0.2\,m in the laboratory, suggesting that the VLM can infer environmental scale from visual context.

\paragraph{Reasoning mode selection guidelines}
No single reasoning mode dominates, but each has a well-defined operating regime visible only when rendezvous and search-and-approach are reported separately. For rendezvous where the target is visible, Standard achieves 13/15 or 87\% with 100\% success at C1 and C2; additional reasoning overhead provides no benefit. For search-and-approach where the target is not visible, Standard and ReAct both collapse to 0/5 under C2 and C3, while Prospective is the only mode that succeeds with a search subtotal of 7/15 at 47\%. For inspection, ReAct achieves the highest score under C3 at 42.8 because its rapid termination limits hallucination accumulation, whereas Prospective's two-stage chain compounds errors. In practice, one may default to Standard and switch to Prospective upon detecting search-phase timeouts.

\paragraph{Self-evolution mechanism insights}
The mechanism excels at \emph{error correction} but provides limited value for already-successful behavior. All four improved groups fail or score poorly at baseline, while the two successful baselines show no benefit. The mechanism is therefore most valuable as an automated failure-recovery system. Learning is fast: three groups improve significantly by Round~1 or~2 from a single failed episode. Compared with Voyager~\cite{wang2023voyager}, Reflexion~\cite{shinn2023reflexion}, and ExpeL~\cite{zhao2024expel}, SpaceMind adds structured skill format compatibility, a quality gate with fingerprint deduplication, and operation in safety-critical physical 3D environments.

\paragraph{Limitations and future directions}
Current VLMs possess limited 3D spatial understanding; SpaceMind compensates through skill knowledge and self-evolution, but the agent still occasionally misjudges spatial relationships under degraded conditions. The self-evolution mechanism provides no benefit when the baseline already succeeds, and the task-scope binding constraint prevents cross-task skill transfer, together limiting its scope to failure recovery. The laboratory validation of 18~runs demonstrates architectural feasibility but lacks statistical power for fine-grained claims, and the UE5 simulation does not incorporate orbital dynamics constraints. Future work includes training VLMs for 3D spatial reasoning in space environments, integrating orbital dynamics, and on-orbit deployment of safe agent decisions (e.g., camera exposure and sensor pointing control) where the decision-control-hub role can be validated without risking spacecraft safety.

\section{Conclusions}
\label{sec:conclusions}

This paper presented SpaceMind, a modular and self-evolving embodied vision-language agent framework for autonomous on-orbit servicing. The framework makes four progressive contributions. First, a modular architecture decomposes agent knowledge, tools, and reasoning into independently extensible dimensions through skill modules with dynamic routing, MCP-based tools with configurable profiles, and injectable mode skills, enabling the same codebase to operate across simulation and physical environments with zero modification. Second, three switchable reasoning modes and hierarchical memory provide configurable cognitive depth, with systematic evaluation across 135~runs revealing that no single mode dominates and each excels in different operating regimes. Third, a skill self-evolution mechanism enables autonomous post-deployment improvement through experience-driven knowledge generation; across 30~runs in six groups, four achieve significant positive gains, including complete failure to 100\% recovery and inspection scores improving from 12 to 59, while two reveal the mechanism's boundary conditions. Fourth, closed-loop validation across 192~runs in both UE5 simulation and a physical laboratory provides empirical support for each design decision and yields actionable engineering guidelines for reasoning mode selection and tool configuration in space agent systems.

SpaceMind demonstrates that a well-structured modular agent framework, combined with autonomous knowledge evolution, can serve as a capable decision-control hub for complex, long-horizon space operations, thereby bridging the gap between current VLM capabilities and the demands of autonomous on-orbit servicing.

\section*{Acknowledgements}

This work was supported by the National Key Research and Development Program of China (Grant No. 2024YFB3909300).

\end{document}